\documentclass[preprint,12pt]{elsarticle}

\usepackage{amssymb}
\usepackage{amsmath}
\usepackage{xspace}
\usepackage{booktabs}
\newcommand{\eg}{{\emph{e.g.}}\xspace}

\journal{Computer Vision and Image Understanding}

\begin{document}

\begin{frontmatter}



\title{LatentDreamer: Single Image to High-Quality 3D Object via Latent Features}




\author[label1]{Huanning Dong}
\author[label1]{Yinuo Huang}
\author[label1]{Fan Li\corref{cor1}} 
\author[label1]{Ping Kuang}

\cortext[cor1]{Corresponding author}

\affiliation[label1]{
    organization={University of Electronic Science and Technology of China},
    city={Chengdu},
    country={China}
}

\begin{abstract}
3D assets are essential in the digital age. While automatic 3D generation, such as image-to-3d, has made significant strides in recent years, it often struggles to achieve fast, detailed, and high-fidelity generation simultaneously. In this work, we introduce LatentDreamer, a novel framework for generating 3D objects from single images. The key to our approach is a pre-trained variational autoencoder that maps 3D geometries to latent features, which greatly reducing the difficulty of 3D generation. Starting from latent features, the pipeline of LatentDreamer generates coarse geometries, refined geometries, and realistic textures sequentially. The 3D objects generated by LatentDreamer exhibit high fidelity to the input images, and the entire generation process can be completed within a short time (typically in 70 seconds). Extensive experiments show that with only a small amount of training, LatentDreamer demonstrates competitive performance compared to contemporary approachs.
\end{abstract}




\end{frontmatter}

\begin{figure}[htbp]
    \centering
    \includegraphics[width=1.0\linewidth]{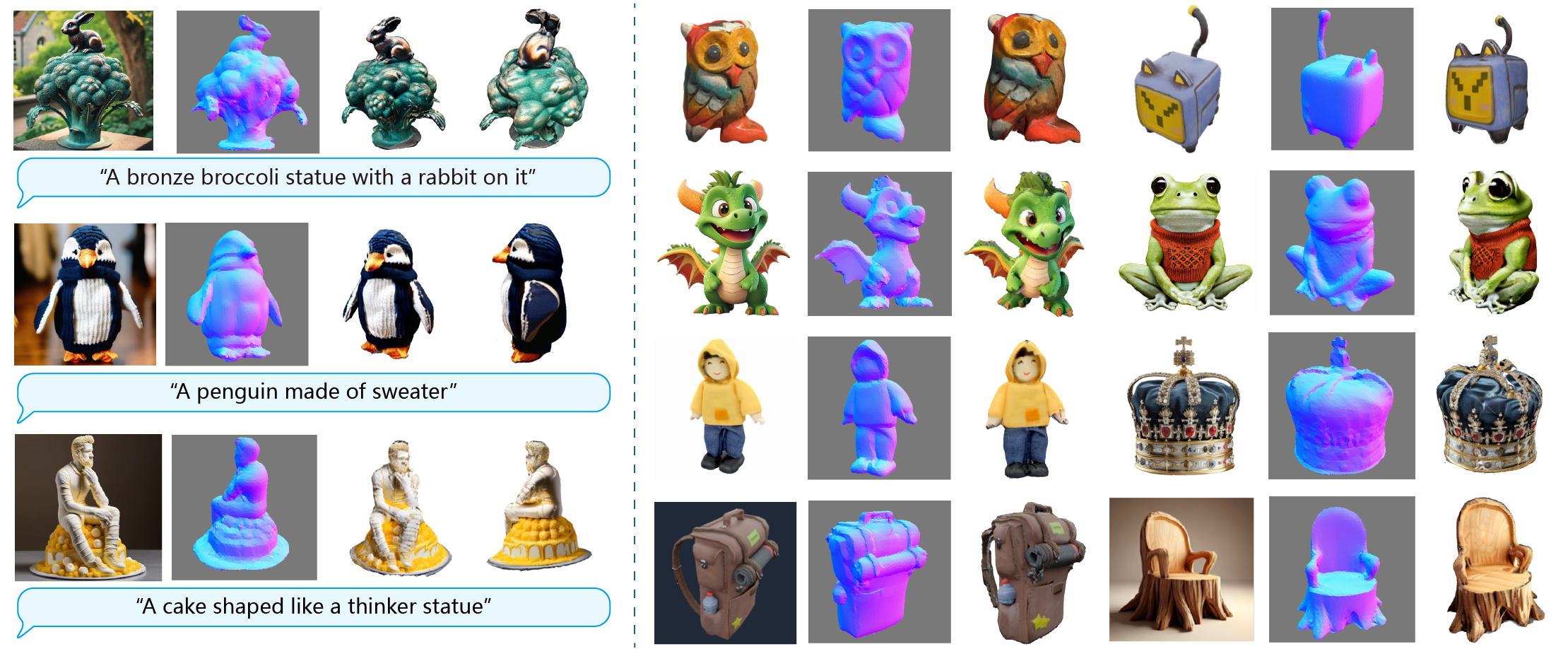} 
    \caption{\textbf{Result of our method.} Our method can generate high-quality textured 3D objects from single images \textbf{in 70 seconds}. Input images (along with text prompts), normal maps of the generated objects, and multi-view color images are shown.}
    \label{fig:headshow}
\end{figure}
\section{Introduction}
\label{sec:intro}

Currently, 3D models are widely used in extensive applications, such as video gaming, animation, education, virtual reality and augmented reality. However, the production of high-fidelity 3D contents often remains a labor-intensive task. Recently, advances in generative AI have shown substantial potential for automatic 3D content generation. For instance, DreamFusion \cite{dreamfusion} employs a pretrained 2D text-to-image diffusion model and the score distillation sampling (SDS) loss to optimize a neural radiance field \cite{nerf} (NeRF). However, such methods are usually time-consuming (DreamFusion takes about 30 minutes to generate a model), as the rendering process of NeRF is slow. Moreover, SDS, as prior knowledge for text-to-image tasks, is prone to the multi-face (Janus) problem due to the lack of explicit image constraints, making it unsuitable for image-to-3D tasks. To address this, some recent studies \cite{zero123,zero123++,wonder3d,syncdreamer,direct2.5,one2345++,mvdream} focus on consistent multi-view images generation that advance image-to-3D development.

Another line of works \cite{crm,sdfusion,3dgen,meshdiffusion,shape,diffusiontosdf,lion} investigate the use of diffusion models \cite{diffusionmodel} to generate 3D content directly. They typically train diffusion models on 3D datasets to generate 3D objects or their intermediate representations. The essence of diffusion-based methods is an iterative denoising process that uses the output of a network (\eg, U-Net \cite{unet}) as a guide for each iteration. Such methods face several inherent challenges, including slow generation speed, lack of details, and low fidelity to input prompts. Concurrent studies, such as LRM \cite{lrm}, explore the potential of transformers \cite{transformers} to perceive 3D space. While these feed-forward methods \cite{one2345,lrm,lgm,eg3d,get3d,instant3d,pflrm} enable faster generation, they depend on extensive training with large amounts of 3D data. However, the limited quantity and variety of public 3D datasets \cite{shapenet,omniobject3d,objaverse} currently hinders their development, leading to subpar generation quality.

In summary, current methods for generating 3D objects from single images have difficulty to balance several challenges: fast generation, accurate geometry, generalization to open categories, and the fidelity to input prompts. In this paper, we propose a novel 3D generation framework that simultaneously addresses the above challenges. First, our method innovatively trains a robust signed distance functions variational autoencoder \cite{vae} (SAE) to encode the discrete representation of the signed distance functions (SDFs) into a global latent features, effectively overcoming the difficulties of optimizing SDFs under sparse views. Notably, during training, we utilize a random geometry generator to create training data, achieving excellent generalization without relying on any public 3D datasets. Second, our approach includes an efficient pipeline for single-image to 3D generation: (1) applying the pre-trained multi-view diffusion model \cite{zero123++} and ControlNet \cite{controlnet} to derive multi-view priors from a single image; (2) training an end-to-end neural network to quickly predict the initialization of latent features; (3) iteratively optimizing latent features to generate coarse meshes; (4) sculpting surface details to create refined meshes; and (5) generating realistic textures for the meshes using view-aware loss. As illustrated in Figure \ref{fig:headshow}, our method can produce high-quality 3D mesh from a single image in just 70 seconds, demonstrating exceptional balance in geometric detail, robustness, generation speed, and multi-category generalization.

In general, our contributions are summarized as follows:

\begin{itemize}
\item We innovatively train SAE that encoding 3D objects into continuous latent features, reducing the difficulty of optimizing 3D objects and enhancing the robustness of generation.
\item We propose an efficient multi-step generation pipeline consists of a self-attention based latent features initialization network (LIN). Our pipeline allows high-quality mesh generation from a single image in 70 seconds.
\item We introduce a random geometry generator when training the SAE and LIN, enabling open-world categories 3D generation without relying on any public 3D datasets.
\end{itemize}

\section{Related Works}
\label{sec:relate}
\subsection{3D generation by optimizing the representation}
\label{sec:3D generation by optimizing representations}
A direct 3D object generation approach is optimize their representations via network inference. Previous researches have explored various types of 3D representations, including meshes, NeRF \cite{nerf}, 3D Gaussians \cite{3Dgaussians}. Certain methods \cite{dreamfusion,magic3d,prolificdreamer,nerdi,dreamfields,it3d,realfusion,latentnerf,dreambooth3d,3dfuse} for optimizing NeRF are prone to problems like floating artifacts or defective geometry in space due to the lack of clear geometric surfaces. In addition, it usually takes tens of minutes or even hours to optimize a NeRF-based model. As an efficient 3D representation, 3D Gaussian \cite{3Dgaussians} offers real-time rendering capabilities. Some studies \cite{gaussiancube,luciddreamer,gsgen3d} focus on optimizing 3D Gaussians to generate 3D objects, achieving faster generation speeds and enhanced visualization. However, the realistic visual quality of the generated objects heavily depends on spherical harmonics. These objects often lack explicit geometric structures and surfaces, which hinders user interaction and presents challenges for further applications in gaming, metaverses, simulations, and other fields. There are also studies \cite{pixel2mesh++,meshdiffusion,fantasia3d,pixel2mesh,text2mesh,clipmesh,magic123} that achieve clear geometry by optimizing meshes (or SDFs). However, these methods are highly dependent on the accuracy of prior knowledge, exhibit poor robustness, struggle with convergence, and are prone to geometric misalignment. In this work, we encode SDFs into continuous latent features, effectively reducing the difficulty of the optimization process.

\subsection{Direct 3D generation}
\label{sec:Direct 3D generation}
Another line of research directly generates 3D objects by training networks. Some studies \cite{crm,sdfusion,3dgen,meshdiffusion,shape,diffusiontosdf,lion} adapting the concept of 2D diffusion and train 3D diffusion models on 3D datasets. For example, SDFusion \cite{sdfusion} and 3DGen \cite{3dgen} train 3D latent diffusion models \cite{ldm} (LDM) to produce 3D objects, but they encounter challenges in generalizing to open-world category objects due to the limitations of the datasets. CRM \cite{crm} trains a diffusion model to produce CCM representations that incorporate geometric information from multi-view images for geometric reconstruction. Other feed-forward methods \cite{one2345,lrm,lgm,eg3d,get3d,instant3d,pflrm}, such as LRM \cite{lrm}, train a large transformer decoder to project image features into a 3D tri-plane representation. LGM \cite{lgm} use an asymmetric U-Net to map multi-view images into multi-view Gaussian features, which are subsequently interpreted and fused into 3D Gaussians. These approaches enhance the speed of 3D object generation, reducing the generation time to under 10 seconds.

We iteratively optimize meshes instead of directly generating them because the results of direct generation highly depend on the priors learned from the datasets. However, 3D datasets remain limited compared to 2D datasets, making it challenging to handle categories that have not been previously encountered. Furthermore, direct generation methods often lack strong constraints from input prompts, leading to lower quality and fidelity of the generated objects.

\subsection{Multi-view Diffusion Models}
There are several fundamental issues associated with using score distillation sampling (SDS) as a prior, including over-smoothing and time inefficiency. LucidDreamer \cite{luciddreamer} identifies the cause of over-smoothing and proposes Interval Score Matching (ISM) to alleviate this issue. But the time-consuming problem still remains.

Recent studies \cite{zero123,zero123++,wonder3d,syncdreamer,direct2.5,one2345++,mvdream} concentrate on generating consistent multi-view 2D images. For instance, Zero123 fine-tunes a large stable diffusion model \cite{stablediffussion} for conditional diffusion generation. Given an image and a target view $(R, t)$, Zero123 predicts the RGB image from the new viewpoint. Zero123 often results in inconsistencies among the generated views. In contrast, Zero123++ \cite{zero123++} generates fixed six-view images, which improves consistency across multiple views. These studies provide accurate pixel-level guidance for optimization-based methods (in Sec \ref{sec:3D generation by optimizing representations}) while effectively solving  multi-face problem.

\section{Methodology}
\label{sec:method}


\begin{figure}
\centering
\includegraphics[width=1\linewidth]{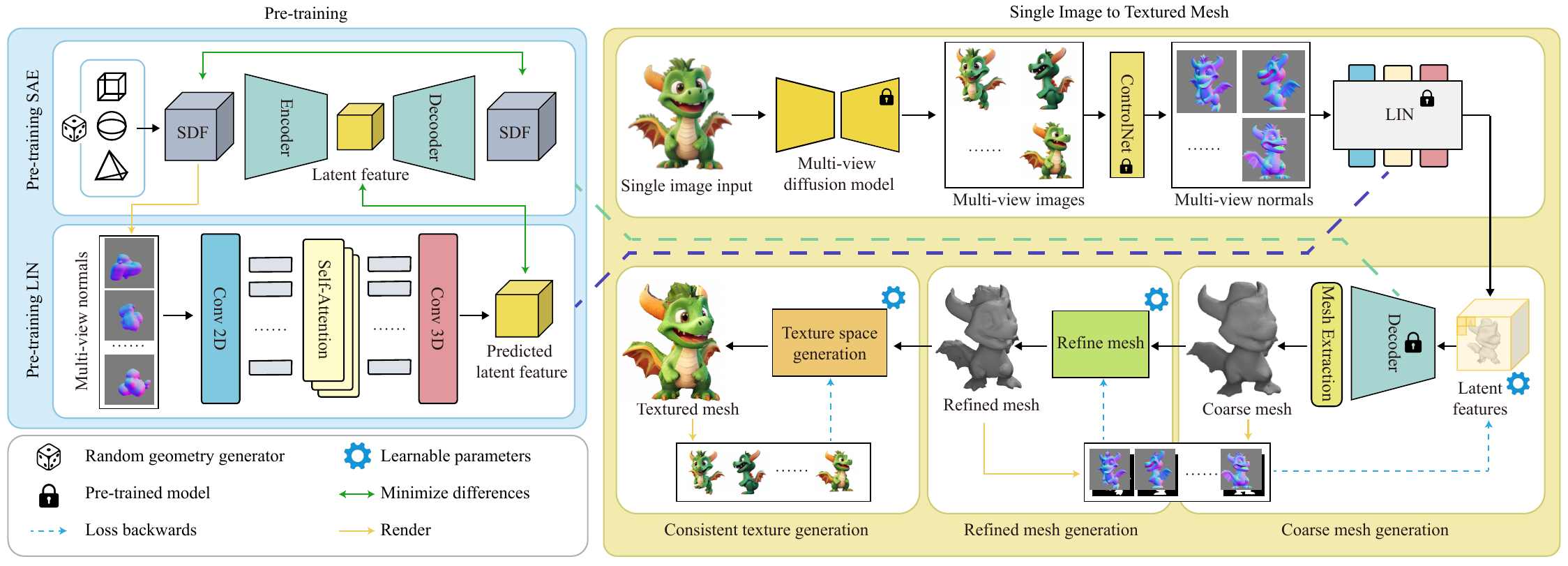}
\caption{\textbf{Overview of LatentDreamer. }\textbf{(Left)} We pre-trained a SAE to decode SDF from latent features. Concurrently, we trained an end-to-end latent feature initialization network (LIN) to predict latent features from multi-view normals for rapid initialization. All training data is generated using a random geometry generator. \textbf{(Right)} For a single image input, LatentDreamer sequentially generates multi-view priors, initialized latent features, coarse mesh, refined mesh, and realistic mesh texture, with the entire process taking only 70 seconds to complete.}
\label{fig:framework}
\end{figure}


In this section, we introduce LatentDreamer. The overview of our framework is shown in Figure \ref{fig:framework}. For single image inputs, we first utilize a pre-trained 2D diffusion model \cite{zero123++} and a ControlNet \cite{controlnet} to generate multi-view priors (in Sec \ref{sec:method Multi-view priors}). Directly optimizing SDFs based on sparse multi-view priors is challenging. We train a SAE that encodes discrete SDFs into continuous latent features, reducing the difficulty of optimizing SDFs (in Sec \ref{sec:method_SAE_for_Any_Geometry}). During mesh generation, we first train an end-to-end latent features initialization network (LIN) to predict the initial values of the latent features (in Sec \ref{sec:method_Initialization_with_Self-Attention}). In Sec \ref{sec:method Single Image to 3D Geometry Generation}, we further employ two rounds of optimization to generate coarse and refined meshes. Finally, in Sec \ref{sec:method Consistent Texture Generation}, we introduce view-aware loss to generate realistic textures for meshes. Notably, we employed a random geometry generator during training to enhance the ability of both the SAE and LIN to handle various geometric shapes.

\subsection{SAE for Any Geometry}
\label{sec:method_SAE_for_Any_Geometry}
Optimizing SDFs directly based on limited 2D guidance can lead to problems like geometric defects and surface holes, as illustrated in Figure \ref{fig:fixation_and_refine} (w/o SDFs fixation). The generated geometry only fits the priors from limited views and lacks of completeness. To address this issue, we pre-train an SAE. The network structure of the SAE is built using 3D ResNet \cite{resnet} layers. We train the encoder to convert discrete SDFs into global latent features. We find that optimizing latent features is easier than directly optimizing SDFs since the decoder is trained to recover latent features into geometry with complete and continuous surfaces. Experiments in Sec \ref{sec:experi} have demonstrated that our approach is effective and robust.

The challenge of obtaining 3D training data is long-standing. We initially attempted to train SAE on public 3D datasets such as ShapeNet \cite{shapenet} and ModelNet \cite{modelnet}. While the model works well within the dataset classes, generating objects of unseen categories presents challenges. Therefore, we propose a random geometry generator to train the SAE using randomly generated 3D geometries, instead of using any public 3D dataset. This strategy achieves impressive results. The decoder can reconstruct nearly any geometry from the latent features, as shown in Sec \ref{sec:experi_analysis}.




\textbf{Random geometry generator.} We assume that each random 3D geometry consists of $n$ random blocks, each of which is described as: 
\[
\text{Block}(S, P, R),\quad
\begin{aligned}
    & S \in \mathbb{R}_{>0}^{3}, \\
    & P \in \mathbb{R}_{>0}^{3}, \\
    & R \in \mathbb{R}_{>0}^{3}
\end{aligned}
\]
Where $S = (L, W, H)$ represents the shape of the block, $P = (x, y, z)$ denotes the coordinate of the block's center, and $R = (\theta_x, \theta_y, \theta_z)$ indicates the rotation of the block around its center. The above parameters $(S, P, R)$ are all randomly generated. A random geometry is represented as:

\begin{equation}
\begin{split}
    Object3d_{0}=\{Block_{i}(S_{i},P_{i},R_{i}) \mid i = 0,1 \dots n\}
\end{split}
\label{eq:objectset}
\end{equation}

To control the roundness of the geometry and the connections between blocks, we apply a random normalized Gaussian kernel to perform $y$ rounds of blurring on each 3D geometry, defined as follows:

\begin{equation}
\begin{split}
    Object3d_{y+1}=F_{blur}(Object3d_{y}), y = 0,1 \dots n-1
\end{split}
\label{eq:objectblur}
\end{equation}

Specifically, in each round of Gaussian blurring, the following calculations are performed for every coordinate in the space:

\begin{equation}
\begin{split}
    SDF'(\boldsymbol {x})=\sum_{\boldsymbol {x'}\in\delta}SDF(\boldsymbol {x'})\cdot G_{normalized}(\boldsymbol {x}-\boldsymbol {x'})
\end{split}
\label{eq:objectgaussion_1}
\end{equation}
where
\begin{equation}
\begin{split}
    G_{normalized}(\boldsymbol {x})=\frac{G(\boldsymbol {x})}{\sum_{\boldsymbol {x'}\in\delta}G(\boldsymbol {x'})}
\end{split}
\label{eq:objectgaussion_2}
\end{equation}
\begin{equation}
\begin{split}
    G(\boldsymbol {x})=\exp\left(-\frac{dist(\boldsymbol {x})^2}{2\sigma^2}\right)
\end{split}
\label{eq:objectgaussion_3}
\end{equation}

Where $\sigma$ is the radius of the Gaussian blur, $\delta$  represents the area within the Gaussian kernel, while dist donates the distance between the coordinate and the center of the Gaussian kernel. $\sigma$ and $y$ are randomly generated in this process. For more details and results about random geometry generator, please refer to our supplement.

\textbf{Loss function:} In this work, we extract meshes from SDFs using Flexicubes \cite{flexicubes}, with its efficient and differentiable mesh extraction ability. Flexicubes defines the loss between the ground truth objects (gt-objs) and the predicted objects (pred-objs) as: 
$\mathcal{L}_{flex}=\mathcal{L}_{render}+\mathcal{L}_{regularization}+\mathcal{L}_{SDF}$. Here, $\mathcal{L}_{render}=\mathcal{L}_{mask} + \mathcal{L}_{depth}$. This component target to rasterize 3D geometry from the same viewpoint and minimize visual differences. $\mathcal{L}_{regularization}$ is defined in Equations 8 and 9 of the paper \cite{flexicubes} and focuses on the internal representation. $\mathcal{L}_{SDF}$ regularizes the geometry by randomly sampling points in both gt-objs and pred-objs and minimizing the distance differences to their surfaces. We incorporated the general losses $\mathcal{L}_{reconstruct}$ and $\mathcal{D}_{KL}$ of VAE. $\mathcal{L}_{reconstruct}$ measures the difference between the inputs and the outputs, here we use $\mathcal{L}_2$ loss. $\mathcal{D}_{KL}$ represents the difference between the latent features distribution and a known distribution, typically expressed as $\mathcal{D}_{KL}(q(\mathbf{z}|\mathbf{x})||p(\mathbf{z}))$.

We discard $\mathcal{L}_{SDF}$ as sampling points and calculating surface distances are time-consuming, significantly reducing training efficiency. We test and find that this operation increases training speed without causing significant performance degradation. Moreover, we add $\mathcal{L}_{normal}$ to $\mathcal{L}_{render}$ to achieve more refined results by comparing the differences in multi-view normals. In general, we define the loss as follows (The coefficients are omitted):

\begin{equation}
\begin{split}
\mathcal{L}_{V}=\mathcal{L}_{render}+\mathcal{L}_{regularization}+\mathcal{L}_{reconstruct}+\mathcal{D}_{KL}
\end{split}
\label{eq:sae loss 01}
\end{equation}
where
\begin{equation}
\begin{split}
\mathcal{L}_{render}=\mathcal{L}_{mask}+\mathcal{L}_{depth}+\mathcal{L}_{normal}
\end{split}
\label{eq:sae loss 02}
\end{equation}

\subsection{Multi-view priors}
\label{sec:method Multi-view priors}

In this work, we utilize Zero123++ \cite{zero123++} to obtain multi-view color images. Given a single input image, Zero123++ employs a pre-trained diffusion model to generate six-view RBG images with uniform azimuths and elevations of 20 and -10 degrees. We then use the pre-trained ControlNet \cite{controlnet} to predict the corresponding normals and further segment them to obtain masks. The masks provide stronger constraints, facilitating the rapid derivation of the approximate 3D geometry. While the priors of six views are sparse for directly optimizing the SDFs, they are sufficient for latent features as discussed in Sec \ref{sec:method_SAE_for_Any_Geometry}. Notably, since our 3D outputs are explicit meshes, it is straightforward to incorporate additional constraints, such as depth information. However, we refrain from using depth data because current monocular depth estimation methods \cite{depthanything,depthanythingv2} do not yield accurate absolute depth information, which could result in incorrect geometry.

\subsection{Initialization with Self-Attention}
\label{sec:method_Initialization_with_Self-Attention}

Through the encoding strategy discussed in Sec \ref{sec:method_SAE_for_Any_Geometry}, we shift the focus of mesh generation from optimizing SDFs to optimizing latent features. However, optimizing geometry from scratch for each input is inefficient. To reduce the convergence time of latent features, we trained an end-to-end latent features initialization network (LIN) that takes multi-view normals as input and predicts the corresponding latent features of the target geometries. As shown in Figure \ref{fig:framework}, LIN first encodes the normals into low-dimensional features and then utilizes self-attention layers to obtain cross-view features. Finally, the latent features corresponding to the normals are predicted. The predicted value will serve as the initial value for the latent feature optimization process.

Similar to SAE, the training data for LIN is also generated by the random geometry generator (in Sec \ref{sec:method_SAE_for_Any_Geometry}), which enhances its generalization ability. LIN is category-independent and focuses solely on the normals without recognizing the categories of the inputs. The loss of LIN is defined as the numerical difference between the predicted latent features and the target latent features (obtained from the encoder of the SAE), rather than the differences observed after rendering. Since latent features with different values may encode similar geometries, we consider that initializations with similar values are more appropriate. Quantitative experiments in Sec \ref{sec:experi ablation} demonstrate the efficacy of our design.

\subsection{Single Image to Refined Mesh Generation}
\label{sec:method Single Image to 3D Geometry Generation}
Although we pre-train SAE to compress SDFs into low-dimensional latent features, our approach does not incorporate the concept of latent diffusion models \cite{ldm} (LDM) to train a diffusion model for generating these latent features, like SDFusion \cite{sdfusion}. Because the latent features generated by the diffusion model lack sufficient accuracy, which diminishes the fidelity between decoded meshes and input images. Instead, we directly optimized the latent features through gradient descent. Specifically, after obtaining the initial latent features $\phi_{0}$ from the multi-view normals through LIN (Sec \ref{sec:method_Initialization_with_Self-Attention}), a coarse-to-fine optimization process is employed to create refined meshes. First, we optimize the latent features iteratively to get coarse mesh, as follows: 

\begin{equation}
\begin{split}
\begin{aligned}\phi^*=\mathop{\arg\min}\limits_{\phi}\mathbb{E}_\theta(R(F(D_{\xi}(\phi), d_0, w_0),\theta)-gt_\theta)\end{aligned}
\end{split}
\label{eq:mesh gen 01}
\end{equation}

Where $D_{\xi}$ represents the decoder of the SAE, which decodes latent features $\phi$ to SDFs. The initial value of $\phi$ is given by the output $\phi_{0}$ from LIN. Flexicubes \cite{flexicubes} takes three parameters: $SDF, Deform$ and $Weight$, and outputs the mesh. $d_{0}$ and $w_{0}$ are the default values of $Deform$ and $Weight$. $R$ denotes the differentiable rasterizer, while $gt$ represents the multi-view priors obtained in Sec \ref{sec:method Multi-view priors}. $\theta$ refers to the camera parameters. After this iteration, we obtain $s^*=D(\phi^*)$, $s^*$ is the approximate optimal value of SDFs which represents the coarse mesh. We then fix $s^*$ and optimize $Deform$ and $Weight$ to carve the surface of mesh. The target is expressed as follows:

\begin{equation}
\begin{split}
\begin{aligned}d^*,w^*=\mathop{\arg\min}\limits_{d,w}\mathbb{E}_\theta(R(F(SDF_{out}, d, w),\theta)-gt_\theta)\end{aligned}
\end{split}
\label{eq:mesh gen 02}
\end{equation}

Overall, we obtain $s^*, d^*, w^*$ as the approximate optimal values of $SDF$, $Deform$ and $Weight$, and further get the refined mesh $Mesh^{*} = F(s^*, d^*, w^*)$. Our experiments in Sec \ref{sec:experi ablation} demonstrate that fixing $s^*$ is crucial to avoid holes and misalignments caused by forced fitting. Throughout the mesh generation process, we render the normals and masks from multiple views and minimize the difference compared to the multi-view priors. The loss is represented as follows:

\begin{equation}
\begin{split}
\mathcal{L}_{G}=\beta \mathcal{L}_{mask}+(1-\beta)\mathcal{L}_{normal}, \beta\in(0,1)
\end{split}
\label{eq:loss mesh}
\end{equation}

We initially set $\beta$ to a high value to quickly obtain a coarse shape, and then gradually reduce $\beta$ to carve the details.

\subsection{Consistent Texture Generation}
\label{sec:method Consistent Texture Generation}

\begin{figure}
    \centering
    \includegraphics[width=0.6\linewidth]{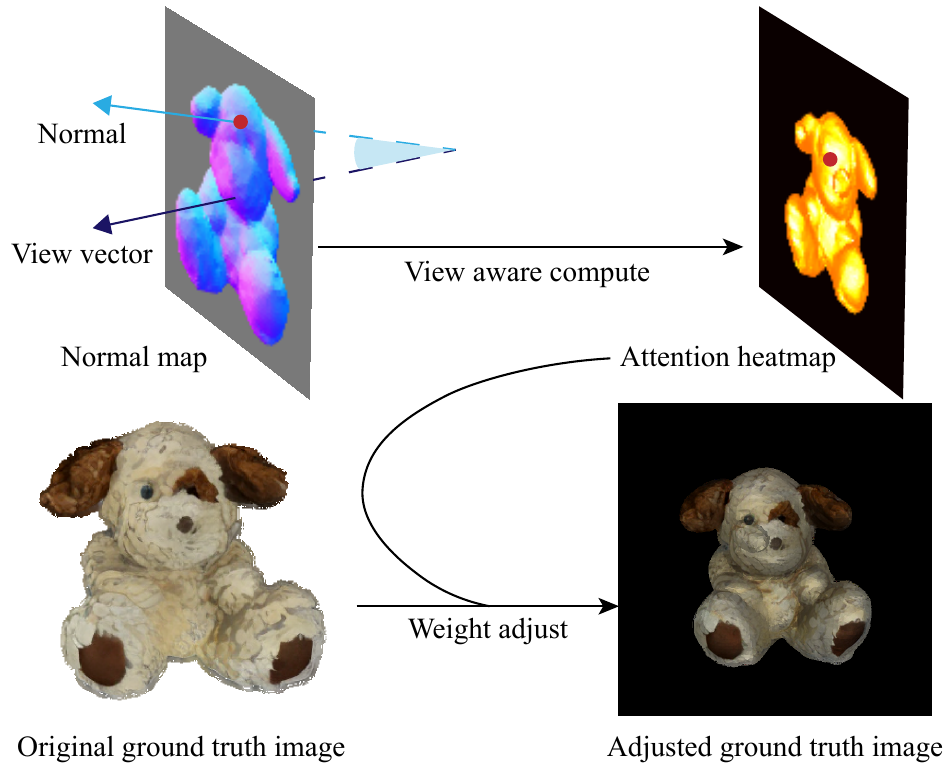}
    \caption{\textbf{View-aware loss.} We assign different weights to pixels based on the direction of their normals. In this figure, brighter points indicate higher weights for those pixels.}
    \label{fig:view_aware_compute}
\end{figure}


\begin{figure}
    \centering
    \includegraphics[width=1.0\linewidth]{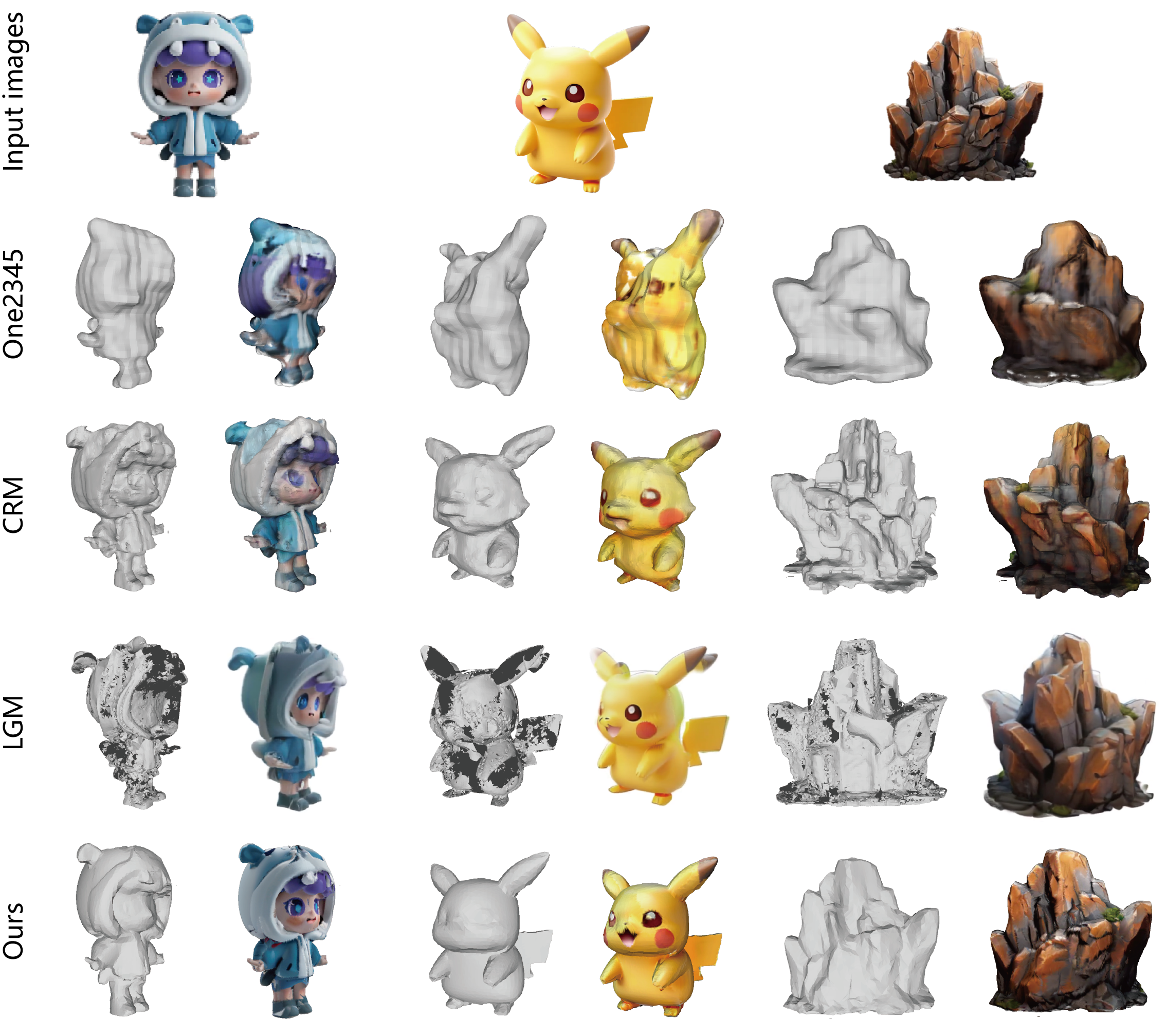}
    \caption{\textbf{Comparison in single-image-to-3D generation.} Input images, meshes, and rendering RGB results are shown. The 3D representation of LGM \cite{lgm} is 3D Gaussians, from where we obtain its rendering results. The other rendering results are derived from the textured mesh. Zoom-in for details.}
    \label{fig:Qualitative_results_01}
\end{figure}


We generate realistic textures by supervising the global texture space ${T(x),x\in \mathbb{Z}^3}$. For each vertex on the mesh, its color is determined by querying the nearest value in the texture space and performing trilinear interpolation. Using differentiable rasterization \cite{diffrast}, we render multi-view RGB images, compare them with the prior multi-view RGB images (in Sec \ref{sec:method Multi-view priors}), and back-propagate the gradient to the texture space. Since the generated prior multi-view RGB images may exhibit inconsistencies, we design a view-aware loss to mitigate potential texture ghosting. The loss is based on the angle between the viewing vector and the surface normal, focusing on the texture loss from critical viewpoints. The general process is illustrated in Figure \ref{fig:view_aware_compute}. With this approach, we obtain realistic and natural textures even in areas where multiple views overlap. The design of our texture loss is as follows:

\begin{equation}
\begin{split}
\mathcal{L}_{t}(x,y)=
\begin{cases}
0,  ~\text{if } \cos (\mathbf{v}_{i}, \mathbf{n}_{x,y}) \leq 0 
\\ (\Delta C_{x,y})^{2} \cdot \cos (\mathbf{v}_{i}, \mathbf{n}_{x,y})^{k}, ~\text{else }
\end{cases}
\end{split}
\label{eq:view aware loss}
\end{equation}

Where $x, y$ are 2D coordinates, $\Delta C$ represents the difference between the prior RGB image and the rendered RGB image, $\mathbf{v}_{i}$ is the current viewing vector, $\mathbf{n}_{x,y}$ is the normal vector at the coordinate, and $k$ is a hyperparameter that adjusts the degree of view awareness.

\section{Experiments}
\label{sec:experi}

\subsection{Implementation Details}
We set the SDFs size to $64^3$ and the texture space to $260^3$. Before generation process, we apply rambg \cite{u2net} for background segmentation to focus on the target object. After mesh generation in Sec \ref{sec:method Single Image to 3D Geometry Generation}, we employ loop subdivision \cite{loopsubdivision} to increase the density of the mesh vertices, allowing for more detailed texture rendering. Notably, the entire training process of LatentDreamer can be completed within two days on a single RTX 3090 GPU without using public 3D datasets, highlighting the efficiency and potential of our method.

\subsection{Qualitative Results}
\label{sec:experi Qualitative Results}

\begin{figure}
    \centering
    \includegraphics[width=1.0\linewidth]{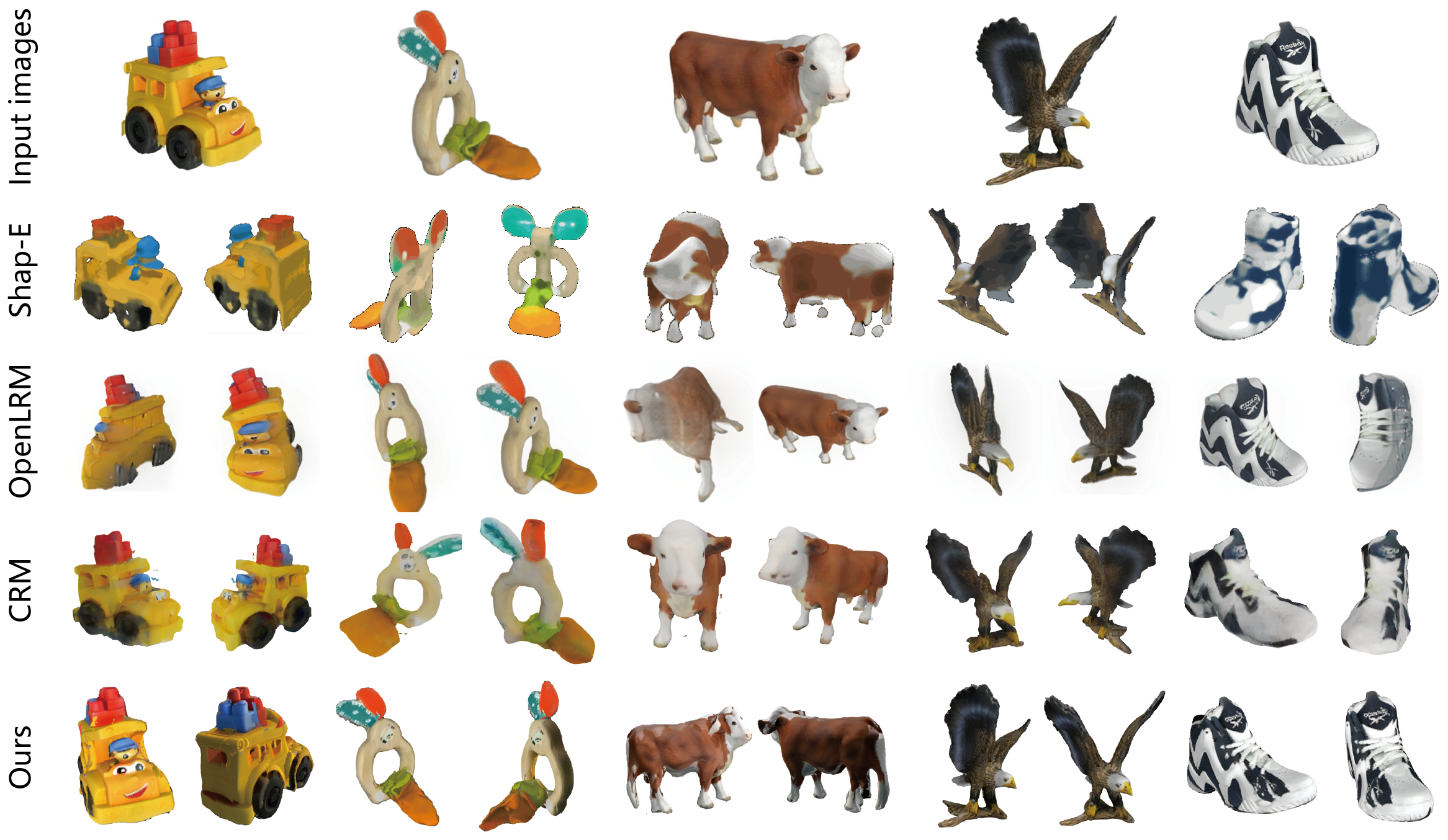}
    \caption{\textbf{Qualitative results on GSO~\cite{gso}.} The 3D objects generated by our method exhibit significantly better fidelity and are free from geometric flattening and distortion.}
    \label{fig:Qualitative_results_02}
\end{figure}

\textbf{Single image to 3D generation. }We compare our method for generating 3D objects from single images with several existing 3D generation methods, including Shap-E \cite{shape}, One2345 \cite{one2345}, LGM \cite{lrm}, CRM \cite{crm}, and OpenLRM \cite{lrm,openlrm}. Qualitative results are presented in Figures \ref{fig:Qualitative_results_01} \ref{fig:Qualitative_results_02}. The examples in Figure \ref{fig:Qualitative_results_02} are sourced from GSO \cite{gso}, providing a general reference. Figure \ref{fig:Qualitative_results_01} displays the untextured mesh and RGB rendering results. However, we only present the multi-view rendering results in Figure \ref{fig:Qualitative_results_02}, since some methods, such as OpenLRM, utilize implicit representations and report only the rendering outcomes.

Among these methods, One2345 can only capture approximate appearances of the objects, resulting in overly smooth outputs. Shap-E can generate 3D objects with styles similar to the input images. However, it often struggles to maintain fidelity to the input images. CRM utilizes a diffusion model to generate intermediate representations, from which it subsequently extracts meshes. However, the resulting meshes exhibit uneven surfaces. LGM employs 3D Gaussians as its 3D representation, which may result in incorrect geometry, such as duplicated Pikachu ears (in Figure \ref{fig:Qualitative_results_01}). OpenLRM use NeRF representation, frequently resulting in distorted geometry and exhibiting impressive visual quality only from the input perspective. In comparison, our method outperforms the above methods in terms of texture detail, fidelity of output results to input images, and overall geometric quality.

\textbf{Text to 3D generation.} In addition to generating high-fidelity 3D objects from single images, LatentDreamer can also be combined with mainstream text-to-image frameworks (\eg Stable Diffusion\cite{stablediffussion}, DALL-E\cite{dalle}) to generate creative and imaginative 3D contents from text descriptions, such as “A penguin made of sweater” in Figure \ref{fig:headshow}.

\subsection{Quantitative Results}

\textbf{Comparison with Baselines. }
We select Point-E \cite{pointe}, Shap-E \cite{shape}, LGM \cite{lrm}, OpenLRM \cite{lrm,openlrm}, and Zero123 \cite{zero123} as baseline methods for quantitative comparisons. Zero123 can generate multi-view images from arbitrary viewpoints based on a single image and can be combined with SDS loss for 3D generation. We utilize the ThreeStudio \cite{threestudio} implementation for Zero123 and the official implementations for the other methods. Our quantitative comparisons are conducted on 200 objects from GSO \cite{gso}. For each object, we render a $512 \times 512$ color image of the generated result and compare it with the input image. The results are presented in Table \ref{tab:compareresult01}, where we report CLIP \cite{clip} similarity, PSNR, SSIM \cite{ssim}, and LPIPS \cite{lpips}. Since  OpenLRM \cite{lrm,openlrm} and Point-E \cite{pointe} also report mesh results, we compare the geometry generation quality with them, as shown in Table \ref{tab:compareresult02}. We align the center of generated meshes to zero, carefully adjust the scale and pose of the meshes. And then calculate Chamfer Distance, Volume IoU and F-score with the ground truth meshes. Both the image rendering results and the geometry quality results show the competitiveness of our method.

\label{sec:experi Quantitative Results}
\begin{table}
\centering
\begin{tabular}{l c c c c}
\toprule
Method & CLIP-Sim$\uparrow$ & PSNR$\uparrow$ & SSIM$\uparrow$ & LPIPS$\downarrow$ \\
\midrule
point-E\cite{pointe}     & 0.956 & 17.950 & 0.893 & 0.127 \\
shap-E\cite{shape}       & 0.872 & 16.758 & 0.892 & 0.189 \\
LGM\cite{lgm}            & 0.963 & 21.420 & 0.910 & 0.078 \\
OpenLRM\cite{lrm,openlrm}& 0.956 & 17.938 & 0.893 & 0.128 \\
Zero123\cite{zero123}    & 0.897 & 18.957 & 0.896 & 0.123 \\
Ours                     & \textbf{0.966} & \textbf{21.515} & \textbf{0.914} & \textbf{0.059} \\
\bottomrule
\end{tabular}
\caption{Quantitative results of the image rendering quality. These results are evaluated on 200 objects in GSO~\cite{gso}. Best values are highlighted in bold.}
\label{tab:compareresult01}
\end{table}

\begin{table}
\centering
\begin{tabular}{l c c c c}
\toprule
Method & Chamfer Dist$\downarrow$ & Volume IoU$\uparrow$ & F-score($\%$)$\uparrow$ \\
\midrule
point-E\cite{pointe}     & 0.0188 & 0.175 & 64.29  \\
OpenLRM\cite{lrm,openlrm}& 0.0176 & 0.316 & 55.08  \\
Ours & \textbf{0.0084} & \textbf{0.525} & \textbf{71.16} \\
\bottomrule
\end{tabular}
\caption{Quantitative results of the geometry generation quality. Among them, we set the resolution for volumetric IoU computation to 128, and the threshold for the F-score to 0.05. Best values are highlighted in bold.}
\label{tab:compareresult02}
\end{table}

\textbf{User study. }We provide single images as inputs and reconstruction results as output, asking users to evaluate the consistency score (cs), geometric quality score (gs), and rendering score (rs). A total of 1380 sets of results were collected from 46 users. The statistical results are presented in Table \ref{tab:userstudy01}, where we report the average ranking and total average scores for the three metrics. Overall, our method is preferred by the majority of users.

\begin{table}
\centering
\begin{tabular}{l c c c c}
\toprule
Method & csRank$\downarrow$ & gsRank$\downarrow$ & rsRank$\downarrow$ & avgScore$\uparrow$ \\
\midrule
shap-E\cite{shape}         & 4.630 & 4.565 & 4.696 & 10.861 \\
One2345\cite{one2345}      & 4.478 & 4.696 & 4.326 & 11.085 \\
LGM\cite{lgm}              & 2.782 & 2.717 & 2.457 & 16.488 \\
OpenLRM\cite{lrm,openlrm}  & 3.935 & 3.696 & 3.783 & 13.429 \\
Ours                       & \textbf{1.630} & \textbf{2.022} & \textbf{2.109} & \textbf{18.631} \\
\bottomrule
\end{tabular}
\caption{\textbf{Statistical results of evaluations from 46 participants.} Higher scores and lower rankings are better. Our method leads in rankings of various metrics and overall performance.}
\label{tab:userstudy01}
\end{table}

\subsection{Analysis}
\label{sec:experi_analysis}

\begin{figure}
    \centering
    \includegraphics[width=0.8\linewidth]{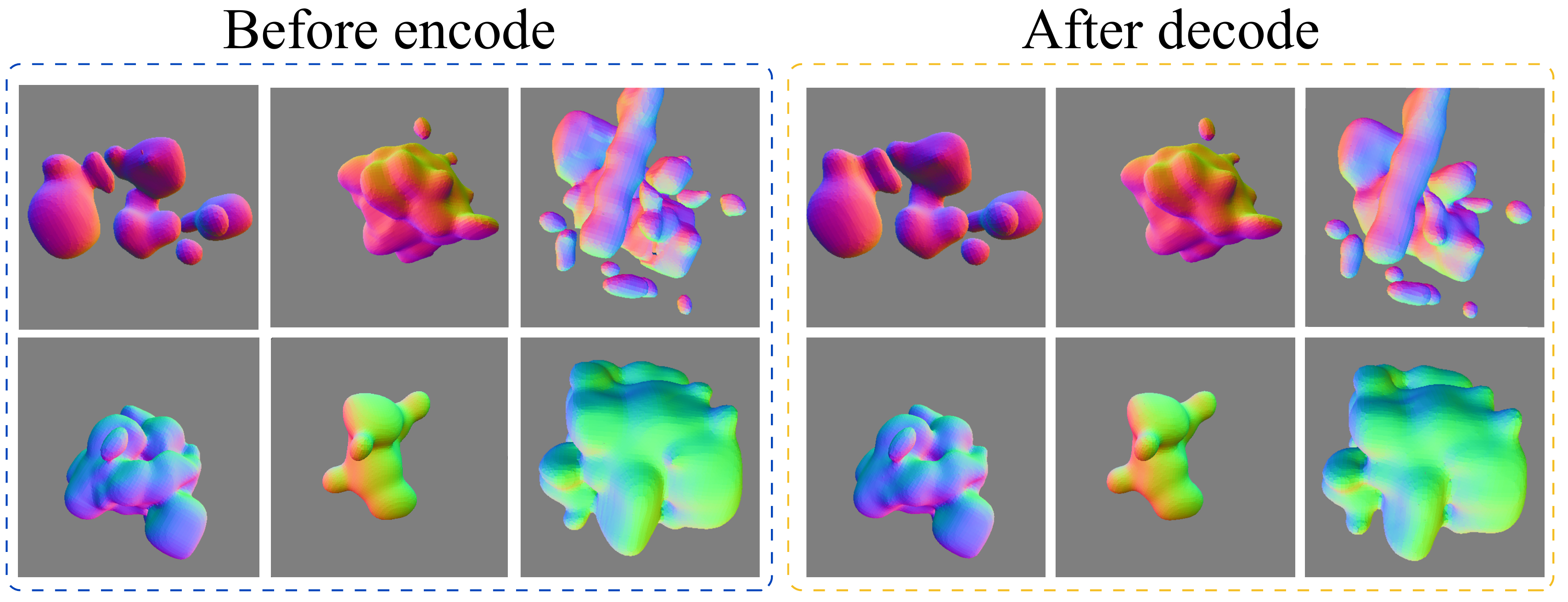}
    \caption{\textbf{Result of SAE.} The left side displays the normals of the input geometries, while the right side shows the normals of the geometries generated by the decoder.}
    \label{fig:result_of_sae}
\end{figure}

\textbf{Reasons for strong multi-category generalization ability. }
Through the SAE described in Sec \ref{sec:method_SAE_for_Any_Geometry}, we encode meshes into latent features that contain a compact amount of data (set to 4096 numbers). These features can represent most geometric shapes well. The experimental results are illustrated in Figure \ref{fig:result_of_sae}.

Our experimental inputs include geometries of varying sizes, multiple separate geometries, and geometries with complex surfaces. The results demonstrate that our encoder–decoder framework effectively compresses and reconstructs arbitrary geometric shapes. The decompressed shapes closely resemble the original inputs while exhibiting smoother surfaces. According to our measurements, the Volume IoU between the input and output geometries can exceed 0.90. The generation of geometric shapes relies entirely on the guidance of latent space optimization and is independent of the object category.

\subsection{Ablation studies}
\label{sec:experi ablation}

Several key strategies are used in LatentDreamer, including latent features initialization, SDFs fixation and mesh refinement. In this section, we verify the effectiveness of these designs through experiments.

\begin{figure}
    \centering
    \includegraphics[width=1.0\linewidth]{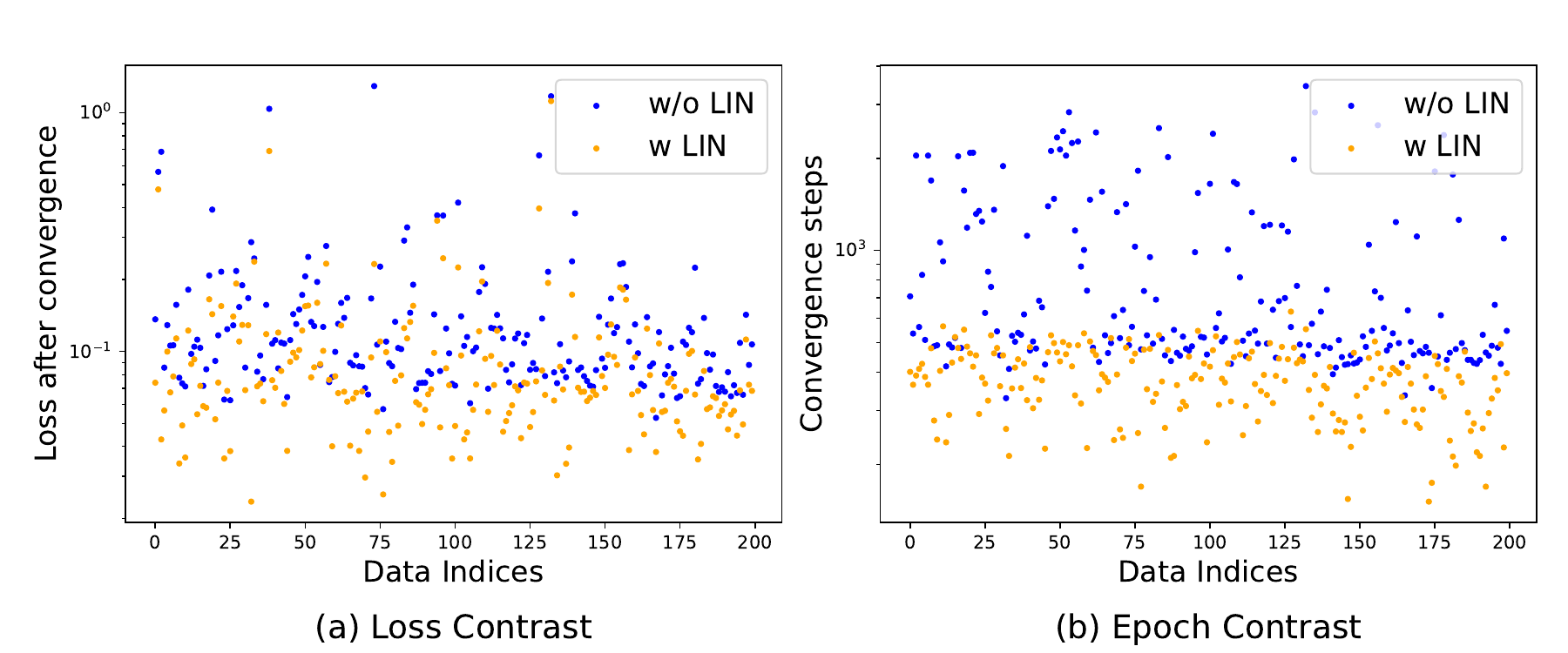}
    \caption{\textbf{Comparison result of whether to use LIN.} (a) The loss after the generation process converges. (b) The number of iterations required for convergence.}
    \label{fig:lin_compare}
\end{figure}

\textbf{Latent Features Initialization Network (LIN). }In this section, we conducted comparative experiments to demonstrate the efficacy of LIN mentioned in Sec \ref{sec:method_Initialization_with_Self-Attention}. The quantitative results are presented in Figure \ref{fig:lin_compare} and Table \ref{tab:LIN}. Our experiments were performed on 200 objects from GSO \cite{gso}.

\begin{table}
\centering
\begin{tabular}{l c c c}
\toprule
LIN & epoch (avg)$\downarrow$ & loss (avg)$\downarrow$ & SD (epoch)$\downarrow$ \\
\midrule
w/o & 884.83 & 0.1462 & 628.08 \\
w/  & 374.03 & 0.0955 & 95.78 \\
\bottomrule
\end{tabular}
\caption{\textbf{Statistical results of whether to use LIN.} Smaller value is better. SD(epoch) measures the stability of the generation process.}
\label{tab:LIN}
\end{table}

Through pre-trained LIN, we can quickly get a suitable initialization of latent features. The convergence time is greatly shortened and the final generation quality is also improved, demonstrating that a suited initialization can prevent the model from getting trapped in a local optimum. Additionally, the gradient descent of the initialized model is more stable, ensuring convergence in a short time.

\begin{figure}
    \centering
    \includegraphics[width=0.8\linewidth]{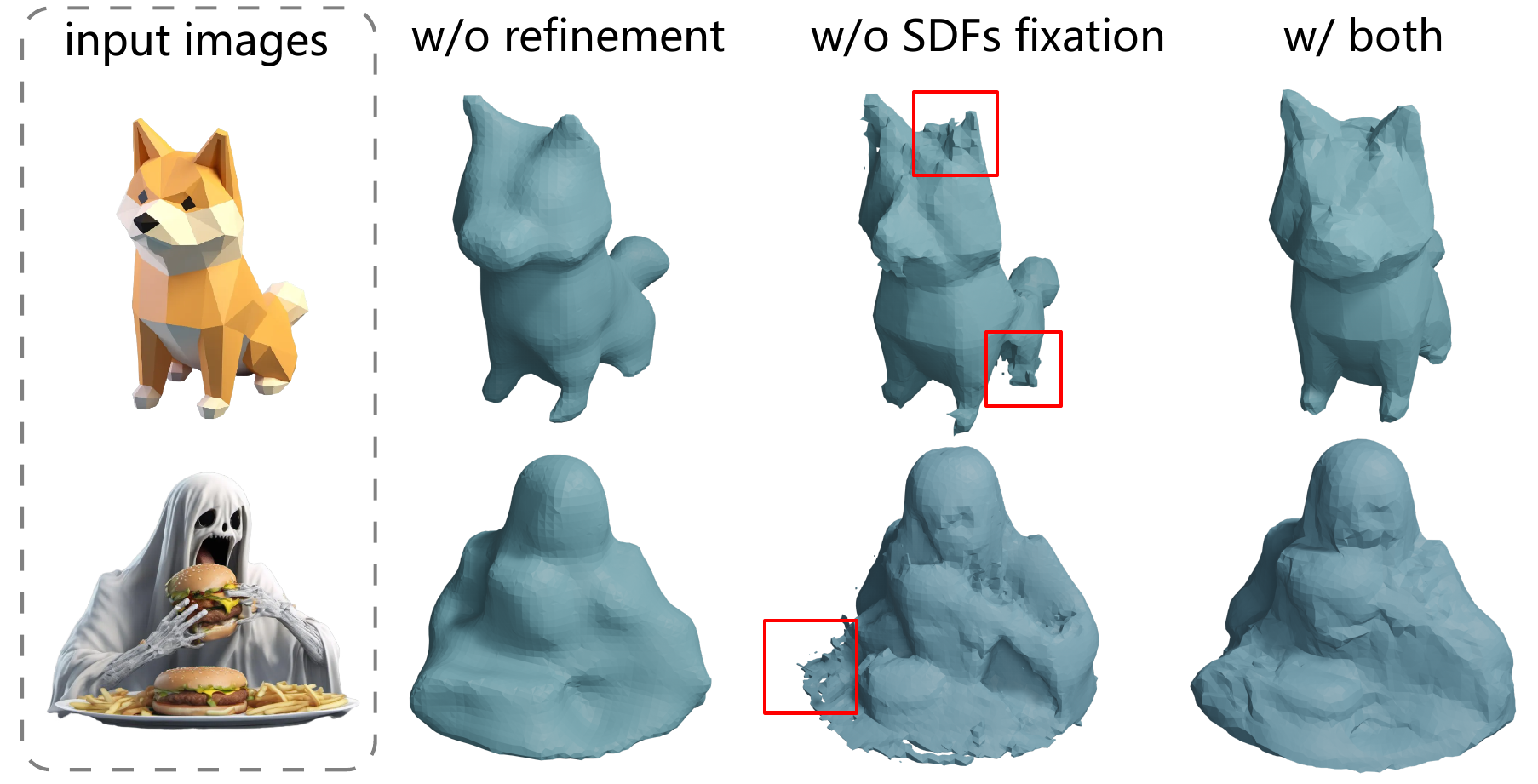}
    \caption{\textbf{Ablation study results of whether to use our strategies.} From left to right: input images, mesh generation without mesh refinement, without SDFs fixation, and with the complete strategy.}
    \label{fig:fixation_and_refine}
\end{figure}


\textbf{SDF Fixation and Mesh Refinement. }
After optimizing latent features to generate coarse meshes, we iteratively refine the $Deform$ and $Weight$ parameters of the Flexicubes \cite{flexicubes} to sculpt geometric surface details while fixing the SDFs to prevent geometric defects (in Sec \ref{sec:method Single Image to 3D Geometry Generation}). Qualitative results are illustrated in Figure \ref{fig:fixation_and_refine}, highlighting the effectiveness and necessity of our design. Our approach enables the generation of complete and detailed geometries.

\section{Conclusions}
\label{sec:conclu}

In this paper, we present LatentDreamer, an innovative method for generating 3D textured meshes from single images. Compared to existing methods, LatentDreamer offers several advantages, including high-quality geometry with detailed surfaces, realistic textures, strong fidelity to the original input image, and efficient generation (typically in 70 seconds). The outstanding performance of LatentDreamer indicates broad application prospects, such as generating 3D objects from personalized text descriptions and controllable 3D character animation. Furthermore, the random geometry generator we propose in this paper can provide a viable solution to the persistent challenges of limited 3D data availability and accessibility.

 \bibliographystyle{elsarticle-num} 
 \bibliography{reference.bib}






\end{document}